# Vehicle Classification under Extreme Imbalance: A Comparative Study of Ensemble Learning and CNNs

Abu Hanif Muhammad Syarubany
Korea Advanced Institute of Science & Technology (KAIST)
hanif.syarubany@kaist.ac.kr

*Abstract* – Accurate vehicle type recognition underpins intelligent transportation and logistics, but severe class imbalance in public datasets suppresses performance on rare categories. We curate a 16-class corpus (~47k images) by merging Kaggle, ImageNet, and web-crawled data, and create six balanced variants via SMOTE oversampling and targeted undersampling. Lightweight ensembles, such as Random Forest, AdaBoost, and a soft-voting combiner built on MobileNet-V2 features are benchmarked against a configurable ResNet-style CNN trained with strong augmentation and label smoothing. The best ensemble (SMOTE-combined) attains 74.8% test accuracy, while the CNN achieves 79.19% on the full test set and 81.25% on an unseen inference batch, confirming the advantage of deep models. Nonetheless, the most under-represented class (Barge) remains a failure mode, highlighting the limits of rebalancing alone. Results suggest prioritizing additional minority-class collection and cost-sensitive objectives (e.g., focal loss) and exploring hybrid ensemble or CNN pipelines to combine interpretability with representational power.

## I. Introduction

Accurate identification of vehicle types is essential for intelligent transportation and logistics systems, yet public datasets are typically skewed toward common classes, making rare categories hard to recognize [1]. To address this, we fused images from Kaggle [2], ImageNet [3], and a Bing crawler [4] into a 16-class corpus (≈42 k training / 5 k testing) and generated six balanced variants with SMOTE oversampling [5] and targeted undersampling [6]. We then compared light-weight ensemble learners: Random Forest [7], AdaBoost [8], and a soft-voting [9] combiner built on MobileNet-V2 [10] features, with a configurable ResNet-style CNN [11] that scales from 18 to 101 layers and employs label smoothing plus heavy data augmentation. The best ensemble (SMOTE-combined) reached 74.8% test accuracy, while the final checkpoint of CNN achieved 79.19% on the full test set and 81.25% on an unseen EE531 inference batch, confirming that deep models excel overall but still falter on the most under-represented class (*Barge*), underscoring the persistent challenge of extreme imbalance.

## II. Related Work

Earlier vehicle-classification studies used constrained surveillance footage or curated datasets such as CompCars [12] and BoxCars [13], yet all report performance drops on minority classes; remedies include synthetic oversampling, cost-sensitive losses, and focal weighting. Random Forests [7] and AdaBoost [8] are popular under imbalance due to bootstrap aggregation and sample re-weighting, though their recall degrades when classes overlap heavily, prompting hybrid voting schemes to smooth individual errors. Deep CNNs, especially residual networks [11], have set state-of-the-art benchmarks for vehicle recognition, but few works directly compare them with classical ensembles on the same imbalanced data. Our study fills this gap by providing a head-to-head evaluation of tuned ensembles versus a ResNet-inspired CNN on a newly balanced multi-source corpus, showing when lightweight methods suffice and when deeper architectures give clear gains.

## III. Implementation

### III.1 Dataset Preparation

In this project, we use a vehicle image dataset comprising 16 classes. The original dataset, sourced from Kaggle, contained class imbalances, particularly for categories like *Ambulance*, *Barge*, *Cart*, *Helicopter*, and *Segway*. To mitigate this, we augmented the data with images from ImageNet [3] and Bing Web Search [4]. The full dataset composition is shown in Table 1.

Table 1 Overview of training and testing images from three data sources. **Source 1** refers to the original dataset from *Kaggle* [2], **Source 2** contains additional class-balanced images retrieved from ImageNet [3], and **Source 3** consists of web-crawled samples obtained using the *Bing Image Crawler* [4].

| Class | Source 1 Train | Source 1 Test | Source 2 Train | Source 2 Test | Source 3 Train | Source 3 Test |
|---|---|---|---|---|---|---|
| Ambulance | 88 | 44 | 1,300 | 50 | - | - |
| Barge | 160 | 42 | - | - | 270 | - |
| Bicycle | 1,496 | 122 | 1,300 | 50 | - | - |
| Boat | 7,909 | 786 | - | - | - | - |
| Bus | 1,782 | 351 | 1,300 | 50 | - | - |
| Car | 5,390 | 1,391 | - | - | - | - |
| Cart | 22 | 29 | 2,600 | 100 | - | - |
| Helicopter | 517 | 151 | - | - | 232 | - |
| Limousine | 11 | 63 | 1,300 | 50 | - | - |
| Motorcycle | 2,189 | 797 | - | - | - | - |
| Segway | 88 | 65 | - | - | 232 | - |
| Snowmobile | 77 | 46 | 1,300 | 50 | - | - |
| Tank | 121 | 85 | 1,300 | 50 | - | - |
| Taxi | 527 | 221 | 1,300 | 50 | - | - |
| Truck | 1,474 | 559 | 1,300 | 50 | - | - |
| Van | 715 | 396 | 2,459 | 100 | - | - |
| Total | 22,216 | 4,711 | 19,009 | 750 | 734 | - |

In total, we collected 41,959 training and 5,461 testing images. Despite these efforts, some classes, such as *Segway* and *Barge*, remain underrepresented due to data scarcity and Bing API limitations, posing challenges for balanced model training. In the experiments, the training data is split into 80% for the training set and 20% for the validation set.

### III.2 Ensemble Learning Approach

This section presents the implementation of ensemble learning techniques used in the project, specifically Random Forest [7], AdaBoost [8], and Voting Classifier [9]. These models aim to leverage the diversity of multiple learners to improve classification robustness and accuracy.

### III.2.1 Dataset and Preprocessing

MobileNet V2 [10] is used as a feature extractor to generate high-level image representations. In the ensemble learning setup, the final classification layer is removed, retaining only the convolutional backbone to extract feature maps.

To address class imbalance and enable comparative analysis, we applied both oversampling and undersampling techniques. SMOTE [5] was used to boost underrepresented classes, while undersampling reduced the prevalence of dominant ones. These methods produced several training set variations, as shown in Figure 1.

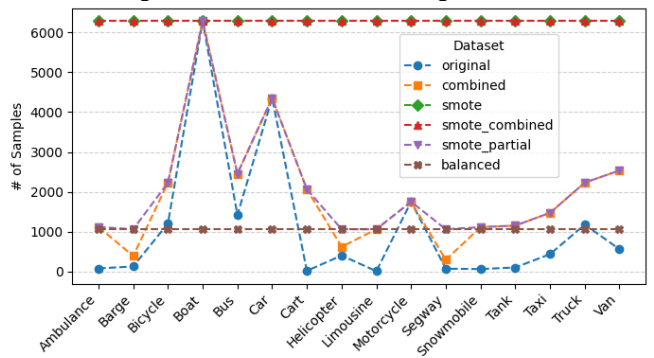

Figure 1 Class distribution across various training set variants. Each line represents original, combined (manual augmentation), SMOTE-based, partially SMOTE-augmented, and balanced (uniform class count).

Figure 1 displays six versions of the training data. The original set shows the raw class distribution, and the combined set includes additional images from external sources. The `smote` and `smote_combined` sets apply full SMOTE to raise all classes to the majority class count. In contrast, `smote_partial` increases only the most underrepresented classes. The `balanced` set further applies undersampling to limit classes exceeding a defined threshold, resulting in a more uniform distribution.

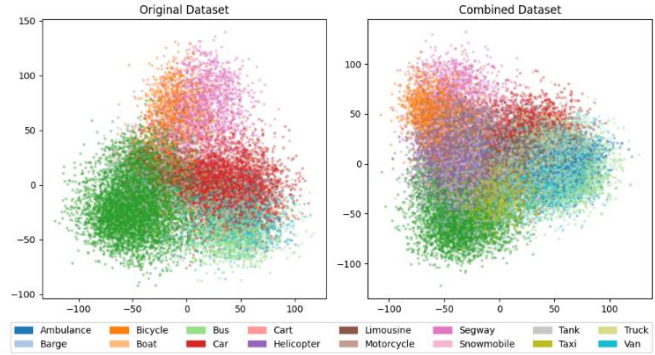

**Figure 2 PCA visualization before applying SMOTE.** PCA projection of the first two principal components from the extracted features of **the original (left)** and **combined (right)** datasets.

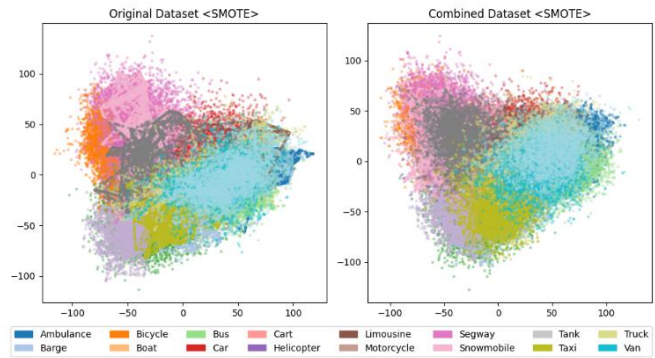

**Figure 3 PCA visualization before applying SMOTE.** PCA projection of the first two principal components from the extracted features of **the original (left)** and **combined (right)** datasets.

Figures 2 and 3 show PCA visualizations of extracted features from the original and combined datasets, before and after SMOTE. In Figure 2, dominant classes such as Car and Bus form dense clusters, while minority classes are sparse. After SMOTE (Figure 3), the clusters become more evenly distributed, indicating improved representation of minority classes in the feature space.

### III.2.2 Model Implementation

In this section, we implement three ensemble models: Random Forest, AdaBoost, and Voting Classifier. Each model undergoes hyperparameter tuning using grid search to optimize performance. The experiments are conducted across six different training set variations to evaluate the impact of data balancing strategies.

### III.2.2.1 Random Forest

In the Random Forest implementation, we first tuned key hyperparameters, specifically the number of estimators and `max_samples`, which controls the proportion of the original data sampled for each tree. After identifying the optimal settings, we evaluated the model's performance across all six training set variations.

### III.2.2.1.1 Parameter Tuning

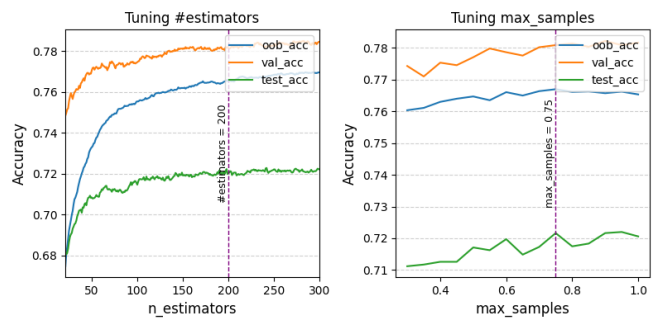

**Figure 4 Hyperparameter tuning results for the Random Forest classifier.** The left plot shows model accuracy as a function of the number of estimators while the right plot shows performance across different proportion of max samples.

The Random Forest model was evaluated using three metrics (as depicted in the Figure 4): out-of-bag (OOB) accuracy, validation accuracy, and test accuracy, while tuning key hyperparameters. As the number of estimators increased, performance improved and stabilized around 200 trees, indicating a good balance between bias and variance. For the `max_samples` parameter, a value of 0.75 provided consistently strong validation and test performance, making it the chosen setting for further experiments.

### III.2.2.1.2 Evaluation (Chosen Model)

Evaluation results across different training dataset variations show notable differences in both validation and test accuracy for original and combined samples (depicted in the Table 2). Among all configurations, the model trained on the `smote_combined` set consistently performs well across both data sources, achieving a strong balance between validation and test accuracy. This suggests that the `smote_combined` variant offers the most robust generalization performance and is therefore selected as the final model configuration.

**Table 2 Evaluation results across training dataset variants.** Results are shown separately for original and combined samples.

| Training Dataset | Original Samples | | Combined Samples | |
|---|---|---|---|---|
| | Val Acc | Test Acc | Val Acc | Test Acc |
| original | 0.8101 | 0.6857 | 0.5643 | 0.6336 |
| combined | 0.8075 | 0.7218 | 0.7808 | 0.7216 |
| smote | 0.8265 | 0.7541 | 0.6461 | 0.7116 |
| smote_combined | 0.8055 | 0.7459 | 0.8008 | 0.7488 |
| smote_partial | 0.8104 | 0.7265 | 0.7794 | 0.7265 |
| balanced | 0.7541 | 0.7108 | 0.7639 | 0.7176 |

The model effectiveness turned out to be coming from the combination of improved class balance via SMOTE and enhanced diversity from external data sources. These aspects help the model learn more representative decision boundaries. This balance is especially important in real-world scenarios where both minority and majority classes must be predicted reliably.

### III.2.2.2 AdaBoost Classifier

In the AdaBoost implementation, we applied grid search to tune key hyperparameters, including the `max_depth` of the base Decision Tree, the learning rate, and the choice of training dataset. This systematic search aimed to identify the optimal configuration that yields the highest classification accuracy across validation and test sets.

### III.2.2.2.1 Grid Search Parameter

**Table 3 Grid Search Parameters Experiments.** Each row represents a configuration group, specifying the training datasets, number of estimators, learning rate values, and the decision tree-based estimator structure

| Experiments | Training Datasets | #Estimators | Learning rate | Estimator $i = max\_depth$ |
|---|---|---|---|---|
| 1st | [original, combined] | [5, 10, 20, 30, 40, 50, 70, 100, 200] | [1e-3, 1e-2, 5e-2, 0.1, 0.2, 0.5, 0.7, 1.0] | Default, $(DT_i)_{i=1}$ |
| 2nd | [original, combined] | [100, 150, 200] | [0.1, 0.2, 0.5] | $(DT_i)_{i=2}^{10}$ |
| 3rd | [ori, smote, combined, smote-combined, smote-partial, balanced] | 100 | [0.1, 0.2, 0.5] | $(DT_i)_{i=2}^{10}$ |

The grid search (as shown in Table 3) for AdaBoost was conducted in three stages, each targeting a different aspect of model optimization. The first experiment focused on identifying suitable values for the number of estimators and learning rate using the default Decision Tree base learner. In the second experiment, the search space was narrowed, and the complexity of the base learner was varied by introducing different `max_depth` values. Finally, the third experiment fixed the model configuration and tested it across various training dataset variants to determine which data distribution enabled the best overall performance.

### III.2.2.2.2 Evaluation (Chosen Model)

**Table 4 Top three configurations from the 1ˢᵗ experiment based on accuracy.** Each row represents a unique combination of training dataset, number of estimators, learning rate, and decision tree depth ($DT_i$).

| Parameters | Original Samples | | Combined Samples | |
|---|---|---|---|---|
| | Val Acc | Test Acc | Val Acc | Test Acc |
| [combined; 100; 0.2; $DT_1$] | 0.72729 | 0.62587 | 0.63081 | 0.60856 |
| [combined; 200; 0.1; $DT_1$] | 0.73637 | 0.63150 | 0.62792 | 0.61169 |
| [combined; 200; 0.2; $DT_1$] | 0.72707 | 0.64666 | 0.65869 | 0.63691 |

Tables 4, 5, and 6 summarize the top three configurations from each stage of the grid search for the AdaBoost classifier. In the first experiment (Table 4), we tuned the number of estimators and learning rate using a default decision stump ($DT_1$). The best-performing configuration used 200 estimators and a learning rate of 0.1, showing moderate accuracy gains. In the second experiment (Table 5), we introduced deeper base learners ($DT_{10}$), which led to improved validation accuracy on the original samples. However, this increase did not translate as effectively to the combined dataset, where performance remained relatively constrained.

**Table 5 Top three configurations from the 2ⁿᵈ experiment based on accuracy.** Each row represents a unique combination of training dataset, number of estimators, learning rate, and decision tree depth ($DT_i$).

| Parameters | Original Samples | | Combined Samples | |
|---|---|---|---|---|
| | Val Acc | Test Acc | Val Acc | Test Acc |
| [ori; 100; 0.1; $DT_{10}$] | 0.77204 | 0.64413 | 0.52235 | 0.59098 |
| [ori; 100; 0.2; $DT_{10}$] | 0.73637 | 0.63150 | 0.62792 | 0.61169 |
| [ori; 100; 0.5; $DT_{10}$] | 0.72707 | 0.64665 | 0.65869 | 0.63691 |

**Table 6 Top three configurations from the 3ʳᵈ experiment based on accuracy.** Each row represents a unique combination of training dataset, number of estimators, learning rate, and decision tree depth ($DT_i$).

| Parameters | Original Samples | | Combined Samples | |
|---|---|---|---|---|
| | Val Acc | Test Acc | Val Acc | Test Acc |
| [smote; 100; 0.5; $DT_9$] | 0.78489 | 0.70687 | 0.56403 | 0.65535 |
| [smote; 100; 0.5; $DT_{10}$] | 0.79840 | 0.71348 | 0.57507 | 0.66423 |
| [smote-combined; 100; 0.5; $DT_{10}$] | 0.76406 | 0.70648 | 0.73968 | 0.70528 |

In the final stage (Table 6), we fixed the model configuration and compared results across different training datasets. Among all variants, models trained on the smote dataset consistently achieved the highest accuracy, with particularly strong performance when using $DT_9$ or $DT_{10}$ and a learning rate of 0.5. These results emphasize the value of balancing class distributions and leveraging moderately deep trees for AdaBoost, contributing to better generalization across both original and combined test sets. Based on these findings, the final model configuration selected for AdaBoost uses the smote-combined dataset, with 100 estimators, a learning rate of 0.5, and a decision tree base learner with max depth 10.

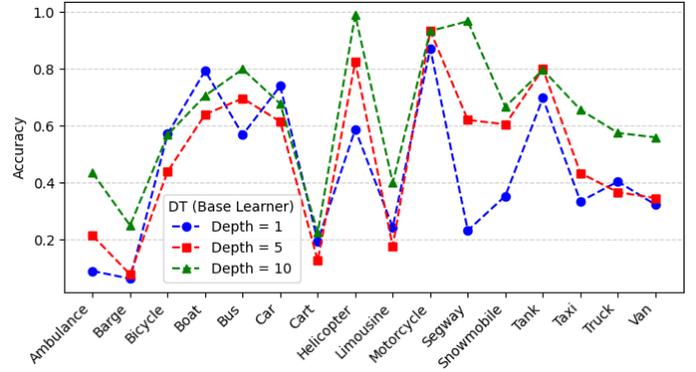

**Figure 5 Accuracy per class.** Achieved by various AdaBoost model with different maximum of depths in the base learner.

To further analyze model behavior, we evaluated class-wise accuracy while varying the maximum depth of the base Decision Tree in the AdaBoost configuration selected earlier, presented in Figure 5. This experiment aimed to assess whether increasing model complexity improves the classifier's ability to distinguish among all classes. As shown in the results, deeper trees ($depth = 10$) generally improve accuracy across most categories. However, AdaBoost still struggles with underrepresented classes such as *Barge, Cart,* and *Limousine*, regardless of tree depth. This limitation stems from AdaBoost's sensitivity to noisy or scarce data, where early misclassifications are heavily weighted and can disproportionately influence subsequent learners, leading to instability when minority classes are insufficiently represented.

### III.2.2.3 Voting Classifier and Performance Conclusion on Esnemble Learning Method

The Voting Classifier is an ensemble method that combines predictions from multiple base learners to improve overall performance. It leverages the strengths of its components by aggregating their outputs, either through majority voting (hard) or averaging predicted probabilities (soft). In this project, we constructed the Voting Classifier using the two best-performing models from earlier experiments: Random Forest and AdaBoost.

### III.2.2.3.1 Performance Evaluation

**Table 7 Evaluation results across ensemble models.** The best-performing models (*Random Forest and AdaBoost*), selected based on prior experiments, are used as base learners in the Voting Classifier.

| Model | Original Samples | | Combined Samples | |
|---|---|---|---|---|
| | Val Acc | Test Acc | Val Acc | Test Acc |
| Random Forest | 0.8055 | 0.7459 | 0.8008 | 0.7488 |
| AdaBoost | 0.7641 | 0.7065 | 0.7397 | 0.7053 |
| Voting Classifier | 0.7922 | 0.7317 | 0.7711 | 0.7333 |

The Voting Classifier yields accuracy values that are generally intermediate between those of its base learners (presented in Table 7). While it does not surpass Random Forest in overall performance, it demonstrates more stable results across both original and combined datasets. This suggests that the ensemble helps mitigate individual model biases and contributes to improved consistency in generalization.

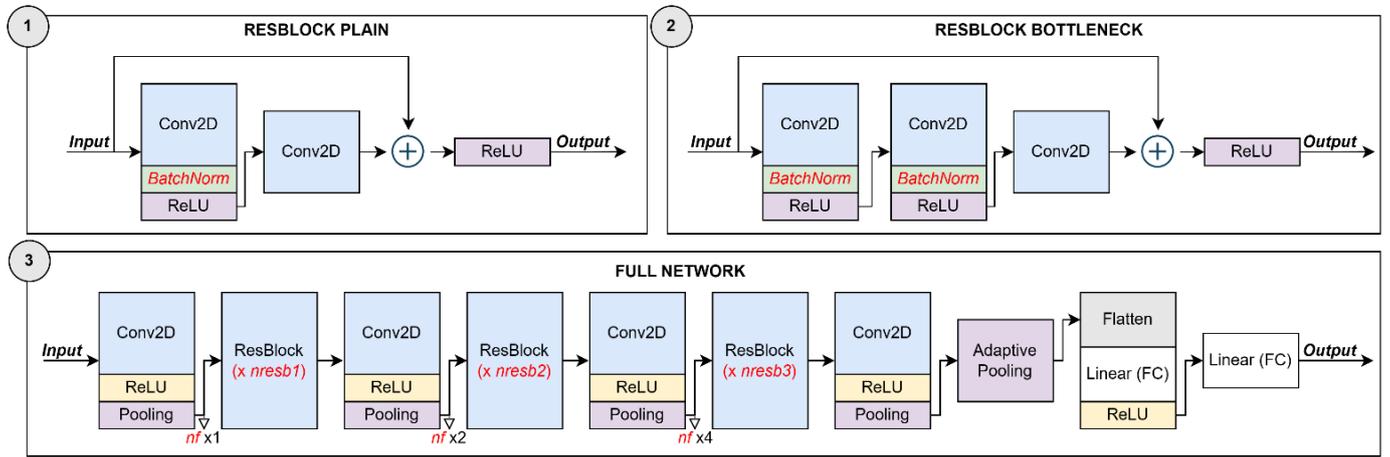

**Figure 9 Architecture overview of the proposed CNN model inspired by ResNet style.** The main network (Block 3) consists of sequential convolutional stages with optional ResBlock modules, which can be configured as either plain (Block 1) or bottleneck-style (Block 2) residual blocks. Red-highlighted variables indicate configurable hyperparameters: number of filters (nf), number of residual blocks (nresb1–3), and the use of Batch Normalization.

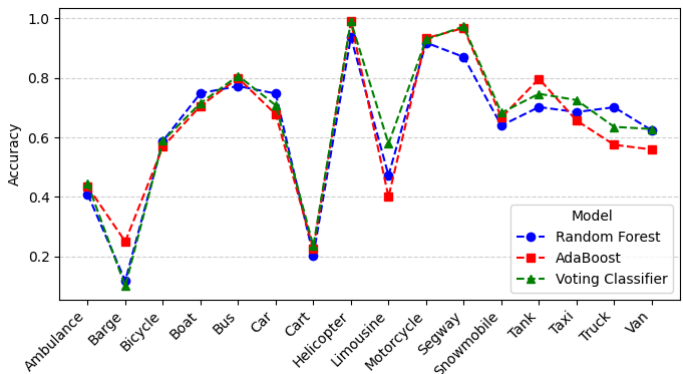

**Figure 6 Accuracy per class.** Achieved by the three ensemble models: Random Forest, AdaBoost, and Voting Classifier.

Figure 6 shows the class-wise accuracy achieved by the three ensemble models: Random Forest, AdaBoost, and Voting Classifier. Across most classes, all models perform comparably, with strong accuracy observed for well-represented categories such as *Motorcycle*, *Snowmobile*, and *Taxi*. However, AdaBoost continues to show weaker performance on several underrepresented classes, such as *Barge*, *Cart*, and *Limousine*. The Voting Classifier generally follows the trends of its base models but demonstrates slightly more consistent performance across classes, highlighting its potential to balance strengths and weaknesses from both Random Forest and AdaBoost.

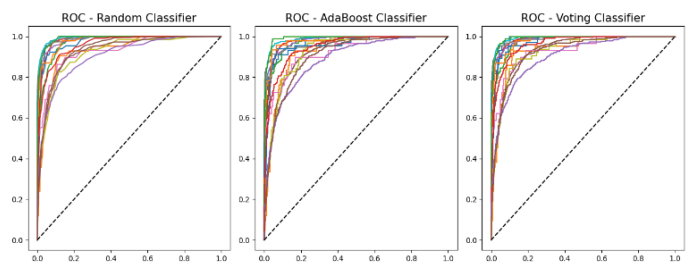

**Figure 7 ROC Curves Comparison.** Represents TPR vs FPR for each class using Random Forest, AdaBoost, and Voting Classifier models.

Figure 7 presents the ROC curves [14] for each class using the Random Forest, AdaBoost, and Voting Classifier models. These plots illustrate the trade-off between true positive rate (TPR) and false positive rate (FPR) across all classes. All three models demonstrate strong discriminatory power, with ROC curves skewed toward the top-left corner, indicating high sensitivity and specificity. Among them, Random Forest and the Voting Classifier exhibit more tightly clustered curves near the ideal region, suggesting more consistent performance across classes. In contrast, AdaBoost shows slightly more variation in ROC shape, which aligns with its earlier observed sensitivity to class imbalance.

III.2.2.3.2 Voting Classifier Inference on Unseen Data

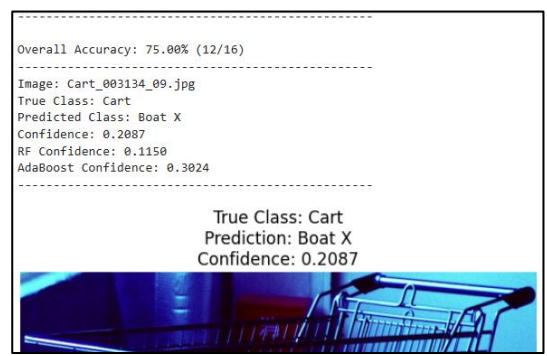

**Figure 8 Inference results on unseen images from the EE531 inference images.** The model (*Voting Classifier*) achieves an overall accuracy of 75% on this set of 16 samples.

To further evaluate its generalization capability, the Voting Classifier was tested on an unseen dataset known as the EE531 inference images, as shown in Figure 8. This dataset consists of 16 samples not previously exposed to the model during training or validation. The Voting Classifier achieved an overall accuracy of 75%, correctly predicting 12 out of 16 images. As shown in Figure 8, some misclassifications still occur, particularly in visually similar classes such as *Cart* and *Boat*, where the model's confidence scores suggest ambiguity. This result highlights the model's strength in handling new inputs, while also revealing the challenge of fine-grained distinctions in visually overlapping classes.

### III.3 Deep Learning Approach

This section outlines the implementation of a deep learning model based on a ResNet-inspired convolutional neural network (CNN) architecture [11]. The experiments focus on designing, training, and evaluating the CNN using various data configurations, aiming to improve performance on multi-class vehicle classification tasks.

III.3.1 Dataset and Preprocessing

The dataset used for training the CNN model combines all three data sources described in Table 1, resulting in a diverse and enriched set of vehicle images. In the initial training phase, preprocessing involved only normalization and resizing to standardize input dimensions. In the second phase, additional data augmentation techniques were applied to improve generalization, including random horizontal flipping, random cropping, color jittering, and random erasing. These augmentations were introduced to simulate real-world variations and reduce the risk of overfitting.

III.3.2 CNN Architecture

The proposed CNN model adopts a modular architecture inspired by ResNet, as illustrated in Figure 9. It consists of an initial convolutional layer followed by three main stages, each composed of a convolution layer and a configurable number of residual blocks. These blocks can be implemented in two forms: *ResBlock Plain*, with two Conv2D layers and optional Batch Normalization, and *ResBlock Bottleneck*, which includes convolution layer sequence to enhance depth while maintaining efficiency.

The network progressively increases its channel depth (from `nf` to `4nf`) and applies pooling after each stage to reduce spatial resolution. The final output is passed through adaptive pooling, then flattened and fed into two fully connected layers to produce the classification logits. The architecture is designed to be flexible, allowing adjustments to the number of filters (`nf`), the number of residual blocks (`nresb1-3`), and the inclusion of Batch Normalization, making it adaptable for various training objectives and dataset complexities.

### III.3.3 Model Implementation

For the model implementation, we conducted two phases of training. The first phase included 18 experiments exploring various configurations, while the second phase extended this by using selected model checkpoints, introducing new deeper architectures, and incorporating data augmentation and label smoothing.

#### III.3.3.1 The First Phase Training Experiment

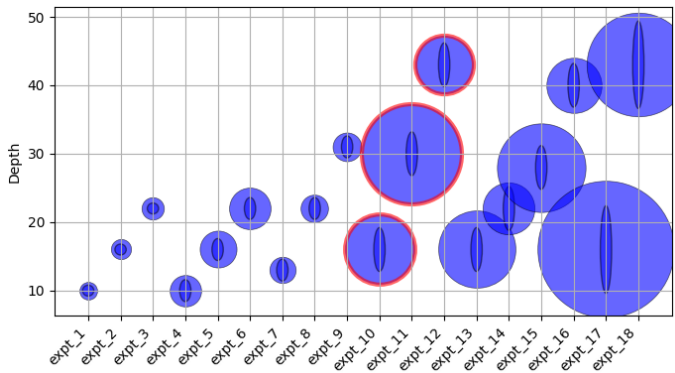

**Figure 10 Visualization of 18 experiments with varying ResNet-style model configurations.** Circle size denotes **# of parameters**, ellipse height reflects **# of filters**, and y-axis indicates **model depth**. Red-outlined circles represent configurations that mimic standard ResNet architectures.

In the first training phase, we conducted a total of 18 experiments using CNN models with varying configurations of depth, number of filters, and residual block designs. The range of configurations is visualized in Figure 10, where each circle represents a different model. Among these, several configurations were intentionally designed to mimic well-known ResNet architectures. Specifically, expt_10, expt_11, and expt_12 correspond to ResNet-18, ResNet-34, and ResNet-50, respectively, and are highlighted with red outlines in the figure.

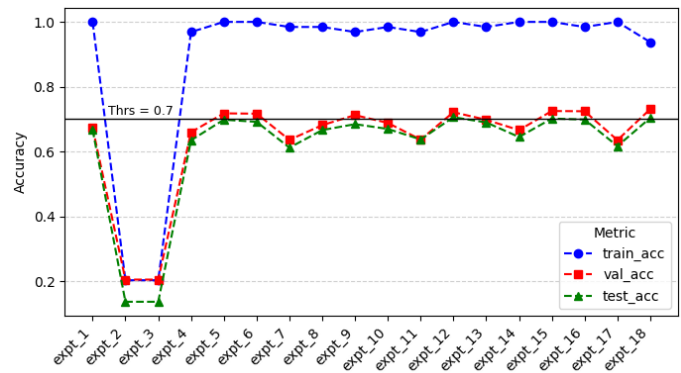

**Figure 11 Evaluation metrics (training, validation, and test accuracy) across 18 experiments.** A 0.7 accuracy threshold highlights models with strong generalization versus those with potential overfitting.

The evaluation results for the 18 training experiments are shown in Figure 11, comparing training, validation, and test accuracies for each model configuration. Most models achieved high training accuracy, but only a subset demonstrated strong generalization as indicated by validation and test accuracy above the 0.7 threshold. Among all experiments, only expt_12, expt_15, and expt_18 surpassed 70% across all three accuracy metrics, indicating reliable performance without overfitting. These three model checkpoints were selected for further training and refinement in the second phase of experimentation.

#### III.3.3.2 The Second Phase Training Experiment

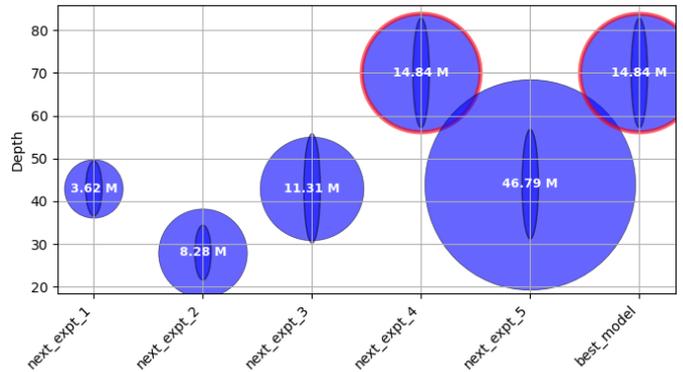

**Figure 12 Extended experiments using data augmentation and label smoothing.** Circle size indicates parameter count, ellipse height reflects filter width, and red-outlined circles represent configurations that mimic standard ResNet architectures.

In the second training phase (as depicted in Figure 12), we extended the experiment by refining the three selected models from the first phase: expt_12, expt_15, and expt_18, which are reused as next_expt_1, next_expt_2, and next_expt_3, respectively. These models were retrained using additional strategies including data augmentation and label smoothing to enhance generalization. The phase also introduced two deeper architectures: next_expt_4 and next_expt_5, with the former mimicking the ResNet-101 structure. Among these, next_expt_4 was further trained with extended epochs, producing the best_model configuration. This best_model checkpoint will later be selected as the final model according to the next section, which has the hyperparameter configurations as follows: $nf = 128$, `bottleneck` type, `batchnorm` ON, and [6,10,6] for the number of 3 residual blocks.

**Table 8 Evaluation results of extended model checkpoints.** Bold values indicate the best performance, while underlined values represent the second best for each metric across train, validation, and test sets.

| Checkpoint | Accuracy | | | Loss | | |
| --- | --- | --- | --- | --- | --- | --- |
| | Train | Val | Test | Train | Val | Test |
| next_expt_1 | 0.8594 | 0.7525 | 0.7284 | 1.0496 | 1.2196 | 0.9031 |
| next_expt_2 | 0.8594 | 0.7415 | 0.7042 | 0.9359 | 1.2394 | 0.9581 |
| next_expt_3 | 0.8125 | 0.7357 | 0.7421 | 1.0302 | 1.2809 | 0.9107 |
| next_expt_4 | 0.8125 | 0.7764 | 0.7590 | 0.9457 | 1.1652 | 0.8122 |
| next_expt_5 | 0.6875 | 0.6572 | 0.6748 | 1.2363 | 1.4345 | 1.0579 |
| best_model | **0.9844** | **0.8188** | **0.7919** | **0.6866** | **1.0877** | **0.7861** |

The evaluation results of the extended training phase are shown in Table 8, comparing accuracy and loss across all checkpoints. Among them, the best_model checkpoint, trained by continuing from next_expt_4 with additional epochs, achieved the highest performance across all accuracy metrics and the lowest training and validation loss. Based on these results, this checkpoint was selected as the final and best-performing model.

### III.3.4 Evaluation (Chosen Model)

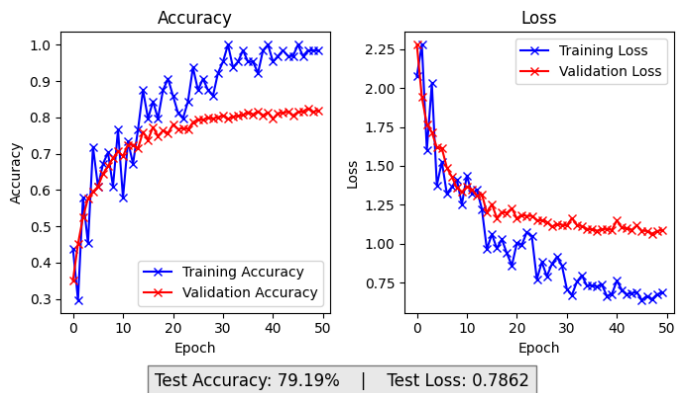

**Figure 13 Learning curves of the best model checkpoint.** The model shows steady improvement with clear convergence and minimal overfitting across 50 training epochs.

The learning progress of the final model shows a clear and steady improvement in both training and validation accuracy, accompanied by a consistent decrease in loss. Throughout the 50 training epochs, the validation metrics remain stable without significant divergence from the training performance, indicating strong convergence and minimal overfitting. This reflects the model's ability to generalize well on unseen data while maintaining effective learning across the training set.

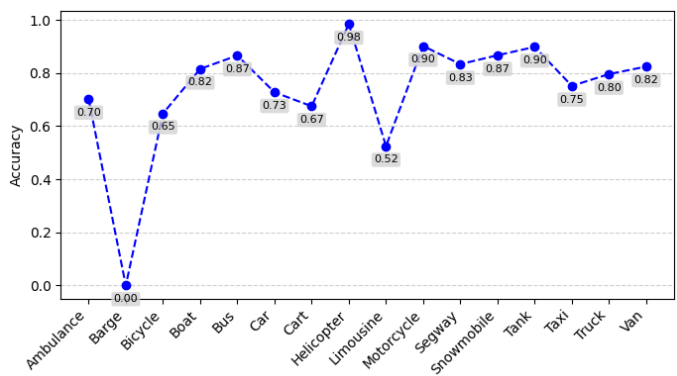

**Figure 14 Accuracy per class achieved by the best model checkpoint.** Highlighting performance variation across vehicle categories.

Although the final model demonstrates strong overall accuracy across most vehicle classes, its performance varies notably by category (as depicted in Figure 14). In particular, the model performs poorly on the *Barge* class, which receives an accuracy of 0.00 despite high scores in other categories. This underperformance aligns with the severe class imbalance observed in the dataset, where *Barge* was among the most underrepresented classes, as shown in Table 1. The result highlights the ongoing challenge of handling extreme data scarcity even in well-optimized deep learning models.

### III.3.5 Model Inference on Unseen Data

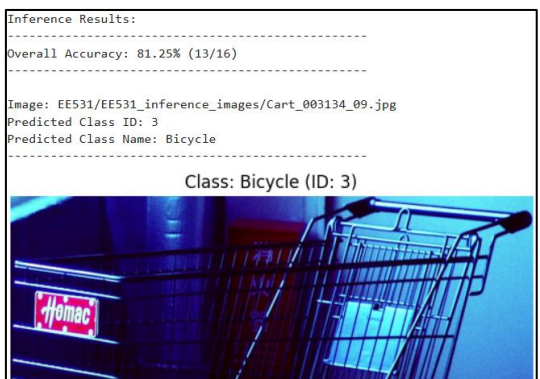

**Figure 15 Inference results on unseen images from the EE531 inference images.** The best model checkpoint achieves an overall accuracy of 81.25% on this set of 16 samples.

To evaluate real-world generalization, the best model checkpoint was tested on the EE531 inference set, which contains 16 previously unseen images (as shown in the Figure 15). The model achieved an overall accuracy of 81.25%, correctly predicting 13 out of 16 samples. This result demonstrates strong transferability and robustness of the trained model, despite some misclassifications in challenging or visually ambiguous cases.

## IV. CONCLUSION

In summary, this study demonstrates that careful data balancing coupled with complementary model strategies can substantially improve multi-class vehicle recognition. Merging three image sources and applying SMOTE-based oversampling and targeted undersampling produced a richer, more uniform dataset that lifted ensemble accuracy to 74.8%, while a deeper, augmentation-enhanced ResNet-style CNN reached 79.19% on the full test set and 81.25% on a completely unseen inference batch. Although the CNN outperformed classical ensembles overall, the poorest results still occurred in the most under-represented class (Barge), underscoring the persistent difficulty of extreme imbalance. These findings suggest that future work should prioritize collecting additional minority-class samples and exploring cost-sensitive or focal-loss [15] training, while also investigating hybrid ensemble–CNN pipelines to combine the interpretability of shallow models with the representational power of deep networks.